\documentclass[10pt,twocolumn]{article}
\usepackage{cvpr}
\usepackage{times}
\usepackage{epsfig}
\usepackage{graphicx}
\usepackage{subfigure}
\usepackage{amsmath}{}
\usepackage{amsfonts}{}
\usepackage{amssymb}
\usepackage{booktabs}
\usepackage{algorithm}
\usepackage{algorithmic}
\usepackage{multirow}
\usepackage{diagbox}
\usepackage{color}
\usepackage{bm}

\usepackage{enumerate}
\usepackage[small]{caption}
\definecolor{aa}{rgb}{0.753,0,0}
\definecolor{bb}{rgb}{0, 0, 1}
\definecolor{cc}{rgb}{0.4275, 0.1686, 0.6039}
\definecolor{dd}{rgb}{0.7020, 0.0667, 0.6627}
\definecolor{ee}{rgb}{0.702, 0.101, 0.929}
\definecolor{hh}{rgb}{0,1,0}
\definecolor{purple}{rgb}{1, 0, 1}

\usepackage{pifont}
\usepackage{overpic}
\usepackage{enumitem}

\usepackage[breaklinks=true,bookmarks=false]{hyperref}

\cvprfinalcopy

\begin{document}

\title{Cross-layer Feature Pyramid Network for Salient Object Detection}

\author{\normalsize{Zun~Li$^{1}$,  Congyan~Lang$^{1}$,
Jun Hao~Liew$^{2}$,
Yidong~Li$^{1}$,
Qibin~Hou$^{2}$,
Jiashi~Feng$^{2}$} \\
	\small{$^{1}$Beijing Jiaotong University, $^{2}$National University of Singapore} \\
	{\small lznus2018@gmail.com,  cylang@bjtu.edu.cn,
	liewjunhao@u.nus.edu, ydli@bjtu.edu.cn,
	{\small andrewhoux@gmail.com,
	elefjia@nus.edu.sg}}
	}
\maketitle

\begin{abstract}
Feature pyramid network (FPN) based models, which fuse the semantics and salient details in a progressive manner, have been proven highly effective in salient object detection.
However, it is observed that these models often generate saliency maps with incomplete object structures or unclear object boundaries, due to  the \emph{indirect} information propagation among distant layers that makes   such fusion structure less effective.
In this work, we propose a novel Cross-layer Feature Pyramid Network (CFPN), in which direct cross-layer communication is enabled to improve the progressive fusion in salient object detection.
Specifically, the proposed network first aggregates multi-scale features from different layers into feature maps that have access to both the high- and low-level information. Then, it distributes the aggregated features to all the involved layers to gain access to richer context.
In this way, the distributed features per layer own both semantics and salient details from all other layers simultaneously,
and suffer reduced loss of important information.
Extensive experimental results over six widely used salient object detection benchmarks and with three popular backbones clearly demonstrate that CFPN can accurately locate fairly complete salient regions and effectively segment the object boundaries. 
\end{abstract}

\section{Introduction}
\label{introSec}
\begin{figure}[!pt]
\begin{center}
\subfigure
{\includegraphics[width=0.45\textwidth]{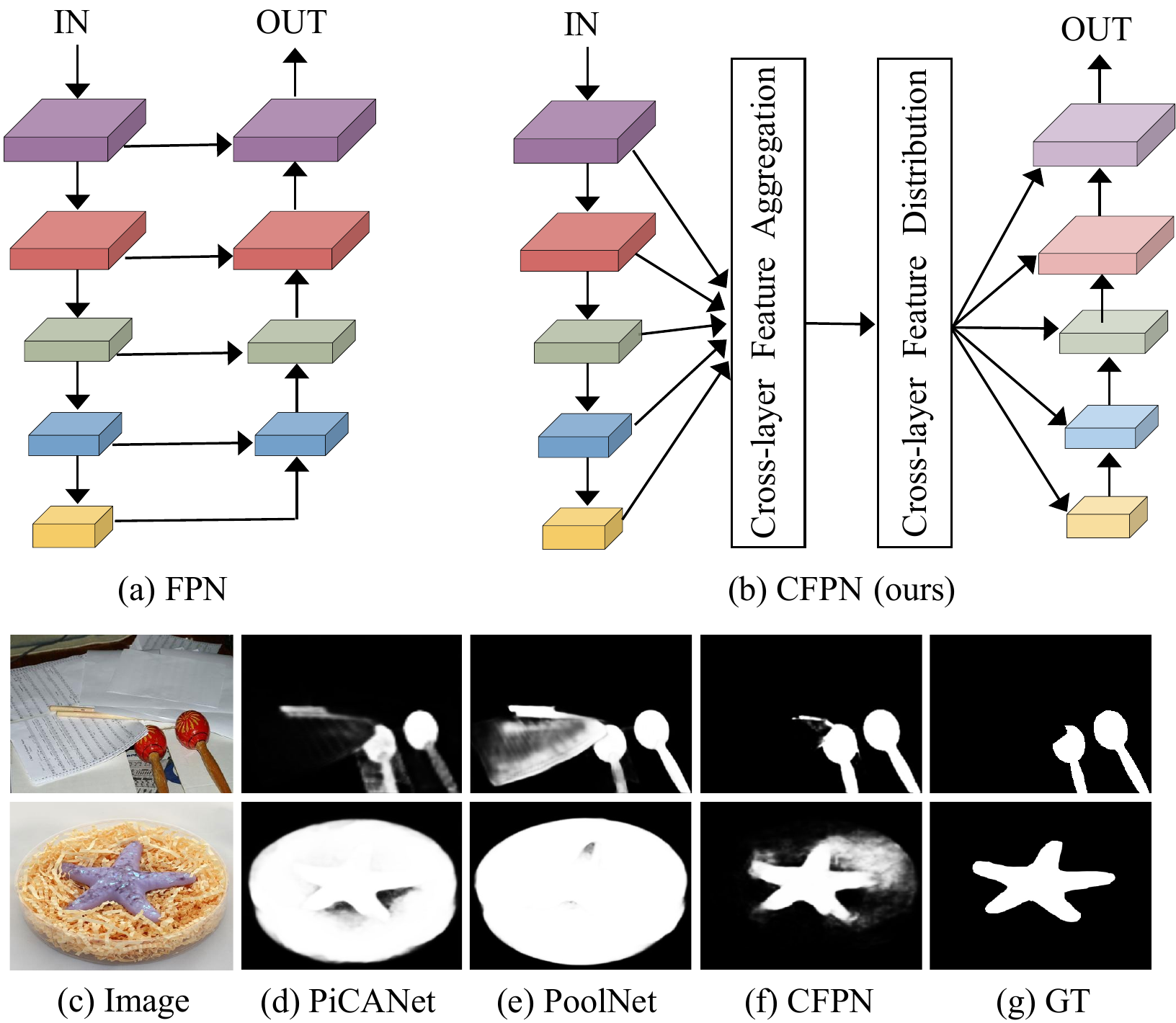}}
\vspace{-4mm}
\captionsetup{margin=1pt,justification=justified}
\caption{Illustration of existing feature pyramid fusion based structure and the proposed CFPN.
Top panel: (a) existing FPN based context fusion structure; (b) pipeline of the proposed cross-layer feature pyramid network (CFPN).
Bottom Panel: (d) and (e) are examples of saliency maps produced by vanilla FPN based saliency methods PiCANet~\cite{PiCANet:liu2018picanet}, PoolNet~\cite{poolnet};
(f) saliency maps generated by our CFPN.
Clearly, saliency maps produced by CFPN show clearer object contour and look closer to the ground truth.
}\label{fig:motivation}
\end{center}
\vspace{-6mm}
\end{figure}
Salient object detection aims to locate and segment the most visually distinctive objects or regions in a given image. 
It serves as a fundamental step in many computer vision tasks like object segmentation~\cite{Wei2015STC,objectSeg}, visual tracking~\cite{tracking,onlinetracking} and photo cropping~\cite{cropping}.
Recently, deep learning based approaches~\cite{DHSNet:liu2016dhsnet,DGRL:wang2018detect,DSS:hou2017deeply,RAS:chen2018eccv,Amulet:zhang2017amulet,SRM:wang2017stagewise,PiCANet:liu2018picanet,poolnet,cascaded,BASNet} have achieved remarkable performance in salient object detection, outperforming the traditional methods~\cite{DRFI,GC,DUT-OMRYang,ECSSDYan} by a large margin. 
Among them,
those leveraging pyramid style fusion~\cite{FPN,unet,hourglass,wu2019mutual}, especially the feature pyramid network (FPN~\cite{FPN}) that progressively fuses multi-scale features in a top-down pathway, have received great attention due to their 
effectiveness for improving localization accuracy and recovering boundary details.

Despite their good performance, 
there is still a large room of improvement for this fusion based approach.
As shown in Fig.~\ref{fig:motivation} (a),
the pyramid fusion structure 
stage-wisely fuses high-level semantics with low-level details via lateral connections.
However,
two drawbacks exist in such approach.
First,
the low-level visual information, such as object edge can only be accessed at the final fusion stage,
{making predicted saliency maps from   those   methods have low-quality   object boundaries.}
Second, as pointed out in~\cite{poolnet,cascaded},
in such pyramid fusion structure,
the high-level semantics are progressively transmitted to the shallower layers,
and hence the semantically salient cues captured by deeper layers may be gradually diluted throughout the progressive fusion.
As a result,
the predicted results 
tend to have incomplete object structures or over-predicted foreground regions.
To alleviate this limitation,
attention models~\cite{PiCANet:liu2018picanet,AFNet,PAGEnet}, gate functions~\cite{gatesaliency,pymaridSal,BMPM:zhang2018bi},
multi-scale feature integration~\cite{SRM:wang2017stagewise,PAGR:zhang2018progressive}, 
and extra supervision (e.g., edge detection \cite{poolnet}, boundary loss \cite{NLDF:luo2017non}) 
have been proposed in the literature. However,
the information propagation is mainly limited between adjacent layers$\footnote{In this paper, layers refer to the side-output features of the backbone.}$ at each fusion stage. Thus, these models still suffer from similar problems
as illustrated in Fig.~\ref{fig:motivation} (d)-(e).

In this work, we propose a novel Cross-layer Feature Pyramid Network (CFPN) aiming at
directly exchanging information across different layers and further boosting information propagation for better salient object detection.
As illustrated in Fig.~\ref{fig:motivation} (b),
CFPN is built on FPN but adopts the following novel architecture designs.
First, it contains a Cross-layer Feature Aggregation module (CFA) that incorporates multi-scale features from different and distant layers to allow communication among different layers.
Among which,
CFA dynamically generates a set of layer-specific aggregation weights to weigh different layer features according to their usefulness for salient object detection.
Second, 
given the reweighted features from CFA,
CFPN also contains a Cross-layer Feature Distribution module (CFD) to allocate the aggregated features to their corresponding layers for the subsequent stage-wise fusion.
Collaborating with CFD,
the distributed features at each layer have access to both semantics and fine details from all other layers simultaneously,
and hence reducing the loss of important information during the progressive fusion.
As a result, better saliency maps can be obtained as shown in
Fig.~\ref{fig:motivation} (f). 
Clearly, benefiting from more direct information propagation among all the layers,
CFPN can predict more complete salient objects with more accurate boundaries.

Our main contributions are summarized as follows:

\begin{itemize}
\vspace{-0.5em}
\item Through analyzing performance limitation of FPN-alike models, we propose that  establishing   direct information communication across multiple layers is important for salient object detection, which has not been considered before. 
\vspace{-0.5em}
 \item We design two novel modules, \ie, the cross-layer feature aggregation module (CFA) and the cross-layer feature distribution module (CFD),  which together allow efficient information communication across multiple layers. 
\vspace{-0.5em}
 \item We develop the  CFPN model based on the above two modules. It can bring consistent performance boost to a variety  of backbones including VGG-16~\cite{VGG}, ResNet-18~\cite{resnet:He2015Deep} and ResNet-50~\cite{resnet:He2015Deep} for salient object detection. It establishes new state-of-the-arts on multiple benchmarks.
 \end{itemize}

\vspace{3mm}
\section{Related Work}
\label{relatedWork}
Early salient object detection methods usually rely on hand-crafted features and heuristic priors~\cite{GC,DUT-OMRYang,DRFI,salientSurvey}, achieving only limited performance due to lack of high-level semantic information.
Recently, benefiting from convolutional neural networks (CNNs), salient object detection enjoys much progress~\cite{DSS:hou2017deeply,HKU-ISLi,Amulet:zhang2017amulet,deepSalientSurvey}.

Some deep saliency methods \cite{HKU-ISLi,DCL:li2016deep,LEGS:wang2015deep,MCDL:zhao2015saliency} divide images into patches or superpixels, and extract single or multiple scales features from each patch or superpixel for determining whether the image regions are salient.
Though better performance has been achieved than traditional methods,
processing images in a patch-wise way ignores the essential spatial information of the whole image, which limits the accuracy for detecting the entire salient objects.

\begin{figure*}[!th]
\begin{center}
\subfigure
{\includegraphics[width=1.0\textwidth]{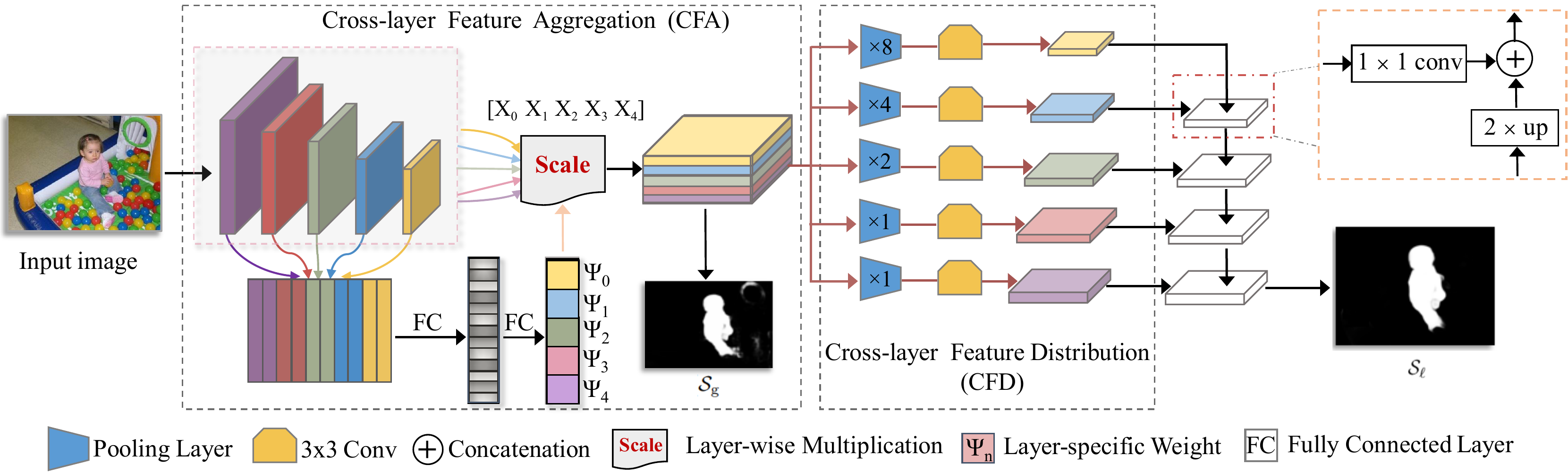}}
\vspace{-5mm}
\captionsetup{margin=8pt,justification=justified}
\caption{Overall framework of CFPN. 
It first extracts local representations ($\mathrm{X_{0}},\mathrm{X_{1}},\mathrm{X_{2}},\mathrm{X_{3}},\mathrm{X_{4}}$) with backbone.
Then, a cross-layer feature aggregation module (CFA) and a cross-layer feature distribution module (CFD) are inserted into the feature pyramid network (FPN) to explore the salient regions. Details of CFA are shown in Fig.~\ref{fig:CAMframework} and Sec.~\ref{CFAsubsection}; details of CFD are presented in Sec.~\ref{CFD}.
}\label{fig:framework}
\end{center}
\vspace{-5mm}
\end{figure*}

Some more effective models are developed based on fully convolutional networks (FCNs)~\cite{FCN:Long}. Wang \etal~\cite{RFCN:wang2016saliency} exploit low-level cues to generate guidance saliency maps by leveraging cascaded FCN.
Liu \etal~\cite{DHSNet:liu2016dhsnet} develop a two-stage network which produces coarse saliency maps first and then integrates local context information to refine them recurrently and hierarchically.
Hou \etal~\cite{DSS:hou2017deeply} introduce short connections into the HED~\cite{HED} architecture, and predict salient objects based on aggregated saliency maps from each side-output.
Wang \etal~\cite{SRM:wang2017stagewise} propose to generate a coarse prediction map via FCN, and then refine it stage-wisely.
Zhang \etal~\cite{Amulet:zhang2017amulet} utilize multi-level context information for accurate salient object detection with the HED network.
In~\cite{DGRL:wang2018detect}, Wang \etal propose to recurrently locate salient objects with local saliency cues.
Zhang \etal~\cite{BMPM:zhang2018bi} extract context-aware multi-level features and utilize a bi-directional gated structure to pass message between them.

Some works introduce the attention mechanism into the network design to exploit multi-level context information for saliency detection.
For example, Zhang \etal~\cite{PiCANet:liu2018picanet} and Liu \etal~\cite{PAGR:zhang2018progressive} both devise attention guided networks in which multiple layer-wise attention is progressively integrated for saliency detection.
Wang \etal~\cite{PAGEnet} first extend regular attention mechanisms with multi-scale information to represent visual saliency contents, and then further improve salient object segmentation performance using salient edge information.

More recently,
the feature pyramid networks (FPNs) \cite{FPN} that are designed in a top-down manner have received growing attention in salient object detection.
Liu \etal~\cite{poolnet} propose a poolnet via plugging topmost level information into FPN fusion branch for detecting the salient objects jointly with the edge detection.
Wu \etal~\cite{cascaded} propose a cascaded partial decoder framework cascading high-level feature maps to
refined the low-level features.
We propose to detect salient objects by conducting cross-layer communication to enhance the progressive fusion of FPN branch for salient object detection.

\section{Method}
\label{CFPN}
\subsection{Overall Architecture}
Fig.~\ref{fig:framework} shows the overall architecture of our Cross-layer Feature Pyramid Network (CFPN).
It consists of two novel components, \ie a Cross-layer Feature Aggregation module (CFA) and a Cross-layer Feature Distribution module (CFD).
The CFA first 
adaptively generates a set of fusion weights
for enhancing the original features at each layer by allowing information 
exchange
among multiple
layers. 
With this, the features are enhanced to have richer contexts.
After CFA, the CFD allocates the aggregated features back to their corresponding layers via multi-scale pooling.
Finally, facilitated by the distributed feature maps, 
CFPN gradually merges them in a top-down manner, similar to FPN, to produce the final saliency output.

\subsection{Cross-layer Feature Aggregation}
\label{CFAsubsection}
\begin{figure}[!t]
\begin{center}
\subfigure
{\includegraphics[width=0.45\textwidth]{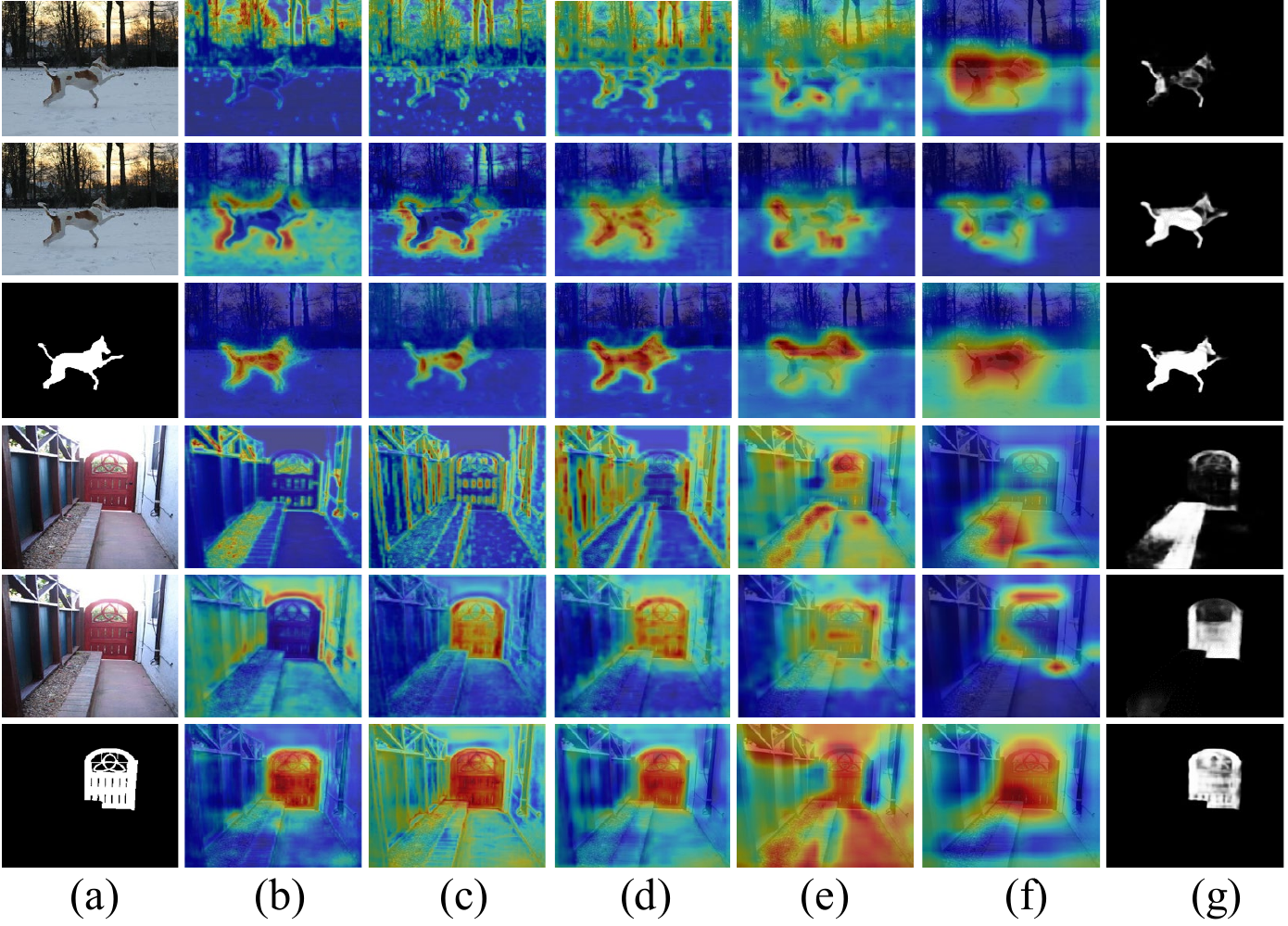}}
\vspace{-4mm}
\captionsetup{margin=4pt,justification=justified}
\caption{(a) Example input images and corresponding ground-truth labels. (b-f) Visualizations of progressive fusion feature maps at different levels from FPN (Row 1, 3), PoolNet~\cite{poolnet} (Row 2, 4), and CFD (Row 3, 6). (g) Saliency maps generated from FPN (Row 1, 3), PoolNet~\cite{poolnet} (Row 2, 4) and CFD (Row 3, 6), respectively.
As can be seen, with our CFD,
feature maps at each level contain richer contexts,
which can more precisely highlight the whole salient objects (Row 3) and effectively suppress the over-predicted foreground regions (Row 6),
compared to the villain FPN based decoder branch (Rows 1, 2, 4, 5). 
}\label{fig:decoder}
\end{center}
\vspace{-4mm}
\end{figure}

As described earlier, FPN based approaches often produce incomplete saliency maps due to gradual dilution of semantics during the progressive fusion.
See
Row 1 and 4 in Fig.~\ref{fig:decoder} for illustration.
Though recent works~\cite{poolnet,cascaded} propose to aggregate the most top layer information into FPN fusion branch, this problem still exists and harms final results, as demonstrated in Column 2 and 5 in Fig.~\ref{fig:decoder}.
In order to enable direct and more efficient communication among different layers, we propose to improve the fusion mechanism in FPN by aggregating all layer features simultaneously. Specifically, since the importance of different layer features largely depends on the image content, we devise a Cross-layer Feature Aggregation (CFA) module to adaptively predict a set of weights according to the importance of each level feature for aggregation.  In this way, the features more useful for salient object detection will be promoted.

Denote the multi-level features output by the first pooling layer and the following four convolutional blocks of the ResNet~\cite{resnet:He2015Deep} backbone as $\mathbf{X}_{n}, n\in\{0,1,2,3,4\}$.
We first append a $1\times1$ convolutional layer at each level for dimension reduction, resulting in features with channel numbers $d_{n}\in \{64, 128, 256, 256, 256\}$.
CFA then applies global average pooling at each level to squeeze its spatial information,
and further concatenates channel-wise statistics from all the levels to integrate local and global contexts {to construct  multi-scale representations}.
Formally, given each level feature $\mathbf{X}_{n} \in \mathbb{R}^{H_{n} \times W_{n} \times d_{n}}$,
CFA calculates the channel-wise global representation $\mathbf{Z}\in \mathbb{R}^{D\times1}$ by
\begin{equation}
\begin{aligned}
\mathbf{Z} &= \mathop{\Big\|}^{N}_{n=1}\boldsymbol{z}_{n}
= \mathop{\Big\|}^{N}_{n=1}\Big\{\frac{1}{H_{n} \times W_{n}}\sum_{i=1}^{H_{n}}\sum_{j=1}^{W_{n}}\mathbf{X}_{n}(i,j)\Big\},
\end{aligned}
\end{equation}
where $\big\|$ is the concatenation function, $D=\sum_{n=1}^{N}d_{n}$ is the channel number of global representation $\mathbf{Z}$. $N$ refers to the overall index of local feature levels, and the pair-wise $(i,j)$ is the spatial coordinate of the feature map at each level.
\begin{figure}[!pt]
\begin{center}
\subfigure
{\includegraphics[width=0.42\textwidth]{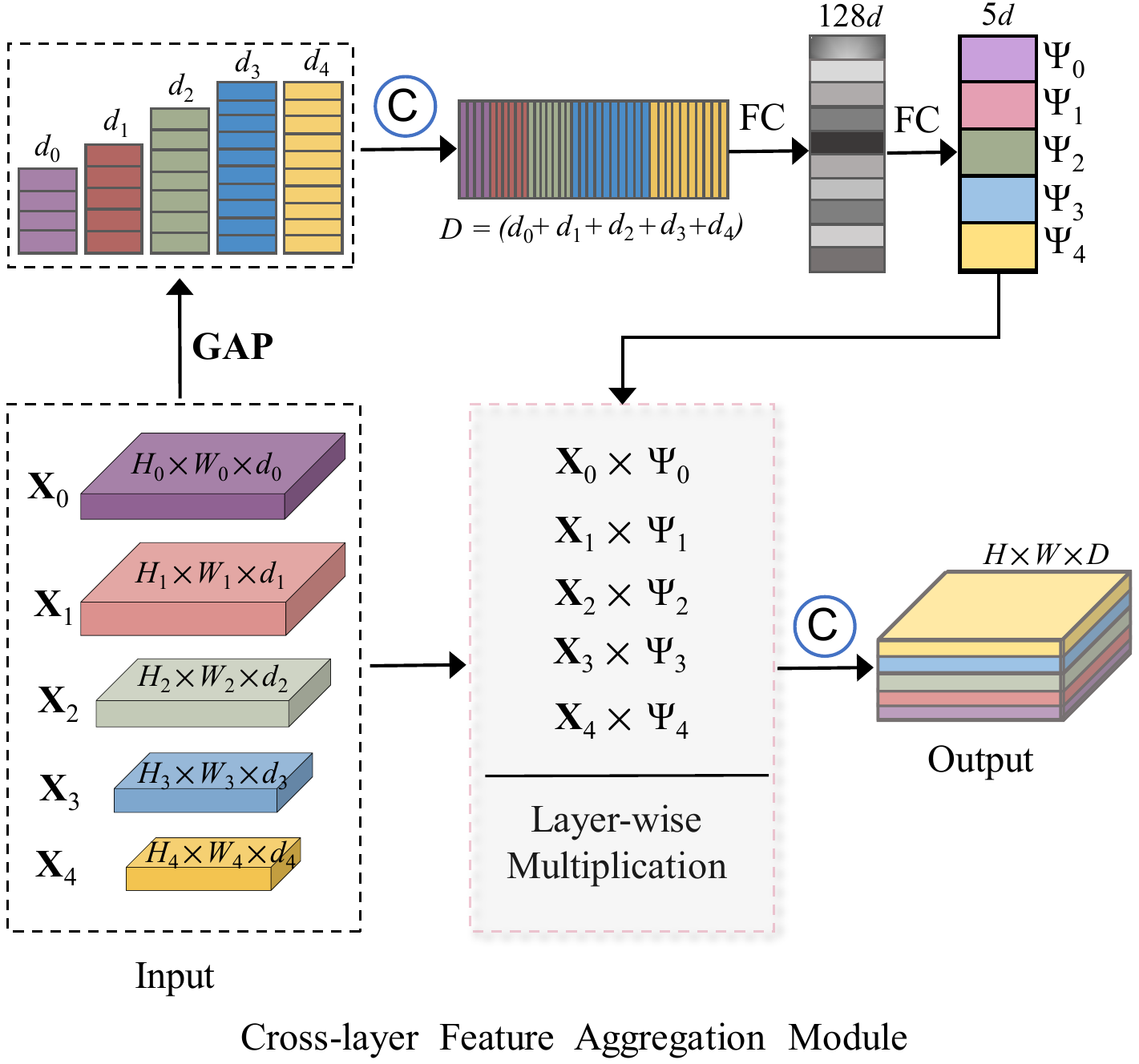}}
\vspace{-3mm}
\captionsetup{margin=8pt,justification=justified}
\caption{Detailed illustration of the proposed cross-layer feature aggregation module (CFA). $\mathrm{\Psi}_{n}, n\in\{0, 1, 2, 3, 4\}$ is the learned layer-wise fusion weight for enhancing features per layer. GAP refers to global average pooling operation. \textcircled{\small{c}} is the feature concatenation operation, and FC refers to the fully connected layer.}\label{fig:CAMframework}
\end{center}
\vspace{-4mm}
\end{figure}

We attempt to leverage the aggregated information $\mathbf{Z}$ to make each level features focus on salient regions instead of the overall feature maps.
To this end, our CFA learns a layer-wise fusion weight $\mathbf{\Psi}\in \mathbb{R}^{1\times N}$ by using a simple gating mechanism for $\mathbf{Z}$, i.e.,
\begin{equation}
\begin{aligned}
\mathbf{\Psi} = \mathbf{W}_{2}\big(\mathrm{ReLU}(\mathbf{W}_{1}(\mathbf{Z}))\big).
\end{aligned}
\end{equation}
Here $\mathbf{W}_{1}\in \mathbb{R}^{D\times M}$ and $\mathbf{W}_{2}\in \mathbb{R}^{M\times N}$ are two fully connected layers inspired by SENet~\cite{senet}, and $M$ denotes the transformed dimension of the global representation, which is set to 128 empirically.
$\mathrm{ReLU}$ denotes the ReLU activation function. With the fusion weight $\mathbf{\Psi}$,
we dynamically enhance each original layer feature by
\begin{equation}
\begin{aligned}
\widetilde{\mathbf{X}}_{n} = \mathbf{X}_{n} * \mathrm{\Psi}_{n},
\end{aligned}
\end{equation}
where $\mathrm{\Psi}_n$ is the $n$-th element in the $\mathbf{\Psi}$,
and $*$ means the scalar multiplication between $\mathbf{X}_{n}$ and $\mathrm{\Psi}_{n}$. 
In this way, the adaptively enhanced multi-level features form
a compact global image representation $\mathbf{F}$ for guiding accurate saliency detection. 
To be more specific,
we first upsample $\widetilde{\mathbf{X}}_{2\sim4}$ to the same resolution as $\widetilde{\mathbf{X}}_0$ by bilinear interpolation,
and then concatenate them to generate the global feature map $\mathbf{F}$. Formally, this process can be expressed as
\begin{equation}
\begin{aligned}
\mathbf{F} = \widetilde{\mathbf{X}}_{0}\oplus 
\widetilde{\mathbf{X}}_{1}\oplus
\mathbf{UP}(\widetilde{\mathbf{X}}_{2})\oplus
\mathbf{UP}(\widetilde{\mathbf{X}}_{3})\oplus
\mathbf{UP}(\widetilde{\mathbf{X}}_{4}),
\end{aligned}
\end{equation}
where $\oplus$ refers to the concatenation operation,
and $\mathbf{UP}$ denotes the upsampling function with bilinear interpolation.

\subsection{Cross-layer Feature Distribution}
\label{CFD} 
Given the aggregated features from the previous CFA module, 
a direct method for producing the saliency map is to convolute the integrated feature with a new convolutional layer.
Although this method can detect salient objects with richer contexts,
the prediction is still not satisfactory 
by using such single stage inference, as shown in Fig.~\ref{fig:CFAglobalmap} (b).
Instead, we propose to combine the aggregated feature maps with FPN, and infer salient regions in a stage-wise fusion manner.
Unlike vanilla FPN, each layer feature now has access to the full spectrum of multi-level representation during the stage-wise fusion, thanks to the aggregation of multi-scale features by the CFA module. Thus, the aforementioned limitations of FPN are largely alleviated.
To this end, 
we devise a Cross-layer Feature Distribution Module (CFD) to allocate multi-level features by performing multi-scale pooling over the aggregated feature $\mathbf{F}$.
In this way, both semantics and salient details can be adaptively accessed at each level of fusion, which boosts the stage-wise fusion in FPN and helps better predict the whole salient objects, as shown in Rows 3 and 6 in Fig.~\ref{fig:decoder}.

Specifically,
CFD first feeds $\mathbf{F}$ to the average pooling layers with pyramid downsampling rates to convert the aggregated features to different scale spaces.
Taking the ResNet version of FPN as an example,
the downsampling rates corresponding to levels $n\in\{0, 1, 2, 3, 4\}$ are
\{1, 1, 2, 4, 8\}, respectively.
Then, a $3\times3$ convolutional layer along with  batch-normalization $(\mathrm{BN})$ and $\mathrm{ReLU}$ activation is appended after each downsampling operation to regenerate feature maps with channel numbers \{64, 128, 256, 256, 256\} as $\widetilde{\mathbf{X}}_{n}, n\in\{0, 1, 2, 3, 4\} $, respectively.
In this way, since the distributed feature maps at each fusion level simultaneously incorporate semantics and fine details, 
more discriminative and complementary representations can be well preserved along the progressive fusion path.
The fusion effect is thus greatly enhanced for achieving more superior performance.

\subsection{Model Training}
 \label{training function}
Given the input image set $\mathcal{I}$ and its corresponding annotations $\mathcal{Y}$,
we train our network with local and global saliency prediction jointly.
This scheme can ensure salient objects uniformly highlighted and backgrounds suppressed, based on our comprehensive experiments.

With the CFA, we obtain the aggregated feature $\mathbf{F}$ with $\mathrm{(\frac{H}{4}, \frac{W}{4})}$ size and $D$ channels.
Then the global saliency map $\mathcal{S}_{\mathrm{g}}$ is predicted with the readout function $\mathcal{R}_{\mathrm{g}}$:
\(\{\mathrm{Conv(3\times3,~128)\to BN\to ReLU\to Conv(1\times1,~1)}\)
\(\mathrm{\to upsampling(H,W)\to sigmoid}\}\). 
\!\!\!\!\!\!
 For predicting the local saliency map $\mathcal{S}_{\mathrm{\ell}}$, after learning the local representation $\mathbf{L}$ from CFD, the prediction function $\mathcal{R}_{\mathrm{\ell}}$: \(\{\mathrm{Conv(1\times1},\!~1)\!\!\!\!\!\to\mathrm{upsampling(H,W)\to sigmoid}\}\), is used to produce $\mathcal{S}_{\mathrm{\ell}}$ directly.
According to $\{\mathcal{S}_{\mathrm{\ell}}, \mathcal{S}_{\mathrm{g}}\}$,
our network is trained by formulating the loss function
\begin{equation}
\begin{aligned}
\mathbb{f} = \mathcal{L}_{bce}(\mathcal{S}_{\mathrm{g}}, \mathcal{Y}|\Theta_{\mathrm{g}})
+ \mathcal{L}_{bce}(\mathcal{S}_{\mathrm{\ell}}, \mathcal{Y}|\Theta_{\mathrm{\ell}}),
\end{aligned}
\end{equation}
where the network parameter $\Theta=\{\Theta_{\mathrm{\ell}}, \Theta_{\mathrm{g}}\}$ is used to generate the saliency maps $\{\mathcal{S}_{\mathrm{\ell}}, \mathcal{S}_{\mathrm{g}}\}$. The $\mathcal{L}_{bce}$ is the balanced binary cross entropy loss
\begin{equation}
\begin{aligned}
\mathcal{L}_{bce}(\Theta) = -\beta\sum_{j\in\mathcal{Y}_{+}}\!\!\mathrm{log}\mathrm{Pr}(\mathcal{Y}^{j}=1|\Theta) \\
- (1-\beta)\sum_{j\in\mathcal{Y}_{-}}\!\!\mathrm{log}\mathrm{Pr}(\mathcal{Y}^{j}=0|\Theta),
\end{aligned}
\end{equation}
where $j$ denotes pixel coordinate, and
$\mathcal{Y}_{+}$, $\mathcal{Y}_{-}$ are the foreground and background label sets, respectively. $\beta$ is the loss weight which is defined as $\beta=\mathcal{Y}_{+}/\mathcal{Y}_{-}$. The salient confidence score $\mathrm{Pr}=(1+e^{-\mathcal{S}})^{-1}$.

\section{Experiments}
\subsection{Settings}
\paragraph{Datasets}
To evaluate the proposed approach,
we experiment on six saliency detection benchmark datasets, including ECSSD \cite{ECSSDYan}, PASCAL-S \cite{PSACALSLi}, DUT-OMRON \cite{DUT-OMRYang},
HKU-IS \cite{MDF:li2015visual}, SOD \cite{SODVida} and DUTS-test \cite{DUTS}, which respectively contain 1,000, 850, 5,168, 4,447, 300 and 5,019 natural complex images with manually labeled pixel-wise ground-truths.

\vspace{-4mm}
\paragraph{Implementation Details}
We perform all experiments using the 
adam \cite{adam} optimizer with initial learning rate 5e-5, 0.9 momentum, 5e-4 weight decay, and batch size 14.
Following previous works~\cite{poolnet,cascaded,PiCANet:liu2018picanet,PAGR:zhang2018progressive,BMPM:zhang2018bi,SRM:wang2017stagewise}, 
we use the training set of DUTS~\cite{DUTS} dataset to train the proposed model. 
The training samples are augmented through random rotation and horizontal flipping.
The backbone (VGG-16~\cite{VGG}, ResNet-18 \cite{resnet:He2015Deep}, and ResNet-50 \cite{resnet:He2015Deep}) parameters of our network 
are initialized with the corresponding models pretrained on ImageNet~\cite{imgnet} and the rest are randomly initialized.
In both training and testing phrases, input images are resized to $384 \times 384$.
Different from some recent saliency models trained with extra supervision constraints (e.g., boundary~\cite{BASNet,DGRL:wang2018detect}, edge~\cite{AFNet,PAGEnet,poolnet}) or post processing operations (e.g., CRF~\cite{DSS:hou2017deeply,PiCANet:liu2018picanet}),
our network simply uses pixel-level saliency annotations, with no extra processes used when generating final saliency maps.

\vspace{-2mm}
\paragraph{Evaluation Metrics}
We adopt three metrics: precision-recall (PR) curves, F-measure, and mean absolute error (MAE) as our evaluation metrics. 
For F-measure,
we report the maximum $F_{\beta}$ (MaxF) for evaluating our method and state-of-the-art approaches, as similar to recent studies~\cite{Amulet:zhang2017amulet,PAGR:zhang2018progressive,DSS:hou2017deeply,
DGRL:wang2018detect,UCF:zhang2017learning,NLDF:luo2017non,R3Net:deng18r,poolnet}.

\subsection{Ablation Studies}
We first analyze the contributions of each module in our method, namely CFA and CFD, to overall performance.
Then, different configurations of feature enhancement strategies are compared to validate our CFA design.
At last, by allocating different numbers of layer features over the aggregartion feature map, we verify the effect of CFD design on improving progressive fusion for detecting salient regions.
All ablation experiments are conducted with ResNet-50 backbone on DUT-OMRON \cite{DUT-OMRYang}, PASCAL-S \cite{PSACALSLi} and DUTS-TE \cite{DUTS} datasets.

\begin{figure}[!t]
\begin{center}
\subfigure
{\includegraphics[width=0.45\textwidth]{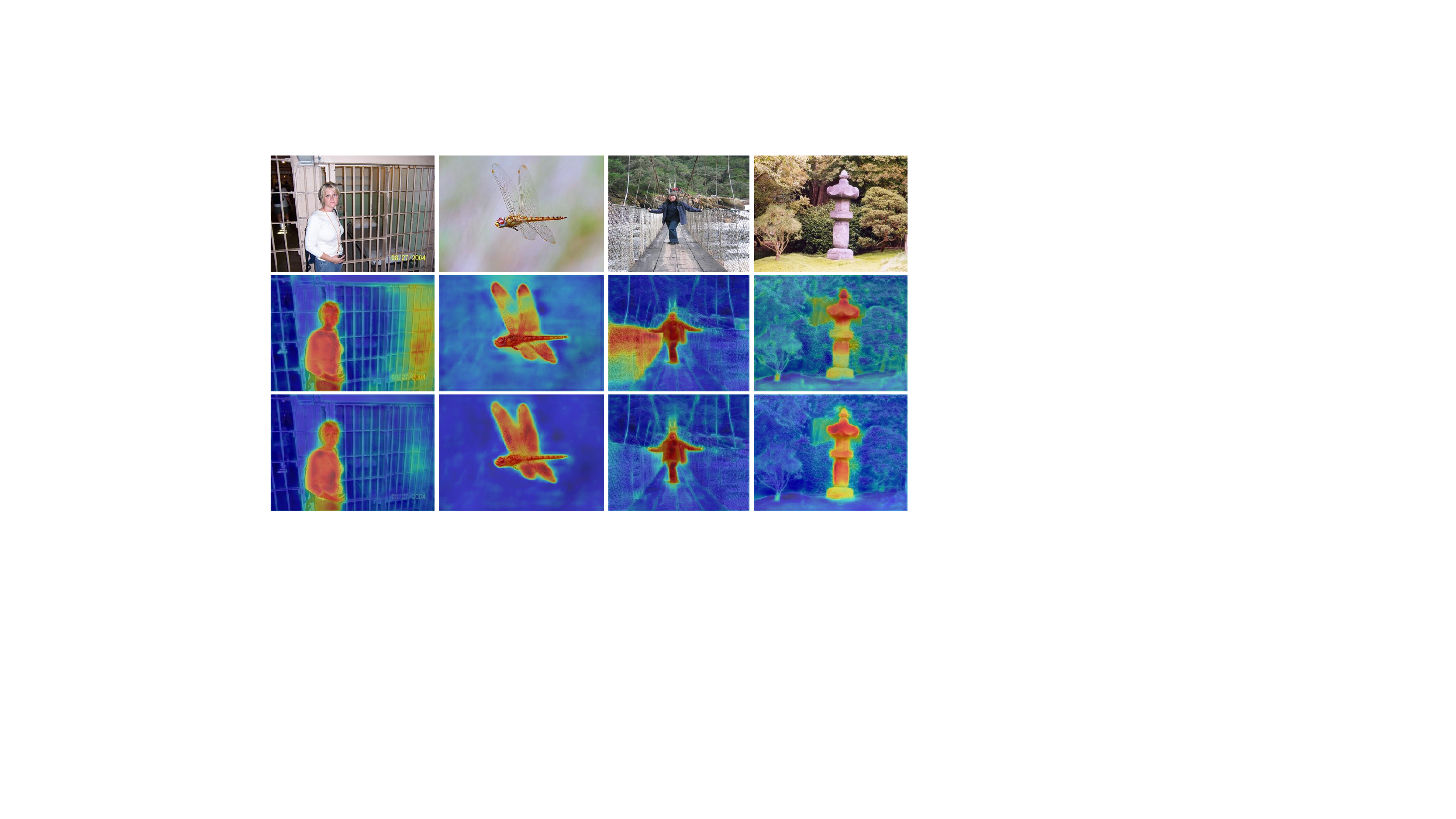}}
\vspace{-3mm}
\captionsetup{margin=8pt,justification=justified}
\caption{Visualizing feature maps generated by directly aggregating the original multiple layer features (Row 2) and
the CFA module (Row 3).
Obviously, feature maps from CFA can more precisely
capture the positions and contours of salient objects (Row 3).
}\label{fig:CFAmaps}
\end{center}
\vspace{-4mm}
\end{figure}

\vspace{-4mm}
\paragraph{Effectiveness of CFA and CFD}
We compare three variants of backbone with the FPN baseline: \textit{w/} CFA, \textit{w/} CFD. Fig.~\ref{fig:decoder}, Fig.~\ref{fig:CFAmaps}, Fig.~\ref{fig:CFAglobalmap} show some visualized results, and Tab.~\ref{component analyse} shows MaxF and MAE scores of CFA and CFD on three challenging datasets.
\begin{itemize}
\setlength{\itemsep}{1pt}
\setlength{\parsep}{1pt}
\setlength{\parskip}{1pt}
 \item \textbf{\textit{w/} CFA}:\quad By comparing results of backbone Res50 (Row 1 in Tab.~\ref{component analyse}, \textbf{\textit{w/o} CFA}),
the addition of CFA (Row 2 in Tab.~\ref{component analyse}) obviously brings performance gain in terms of both MaxF and MAE scores.
Besides,
compared to Row 2 in Tab.~\ref{component analyse},
CFA consistently outperforms the vanilla FPN,
with a margin of 2.1\% and 2.9\%  on DUTS-O, PASCAL-S dataset \wrt MaxF, respectively.
This validates the effectiveness of our dynamic cross-layer feature aggregation strategy.

From visualization results in Fig.~\ref{fig:CFAmaps},
when comparing Row 2 (w/o CFA) and Row 3 (w CFA),
feature maps after CFA provide more discriminative information for distinguishing foregrounds from clutter backgrounds, and thus can better locate the entire salient object than those without CFA.
Moreover, 
by adaptively aggregating multi-layer features,
the CFA greatly improves the quality of generated global saliency maps, as shown in Fig.~\ref{fig:CFAglobalmap} (b).
These results clearly demonstrate that saliency detection benefits from dynamic feature aggregation over information exchanging across multiple layer features.
\begin{figure}[t]
\begin{center}
\subfigure
{\includegraphics[width=0.47\textwidth]{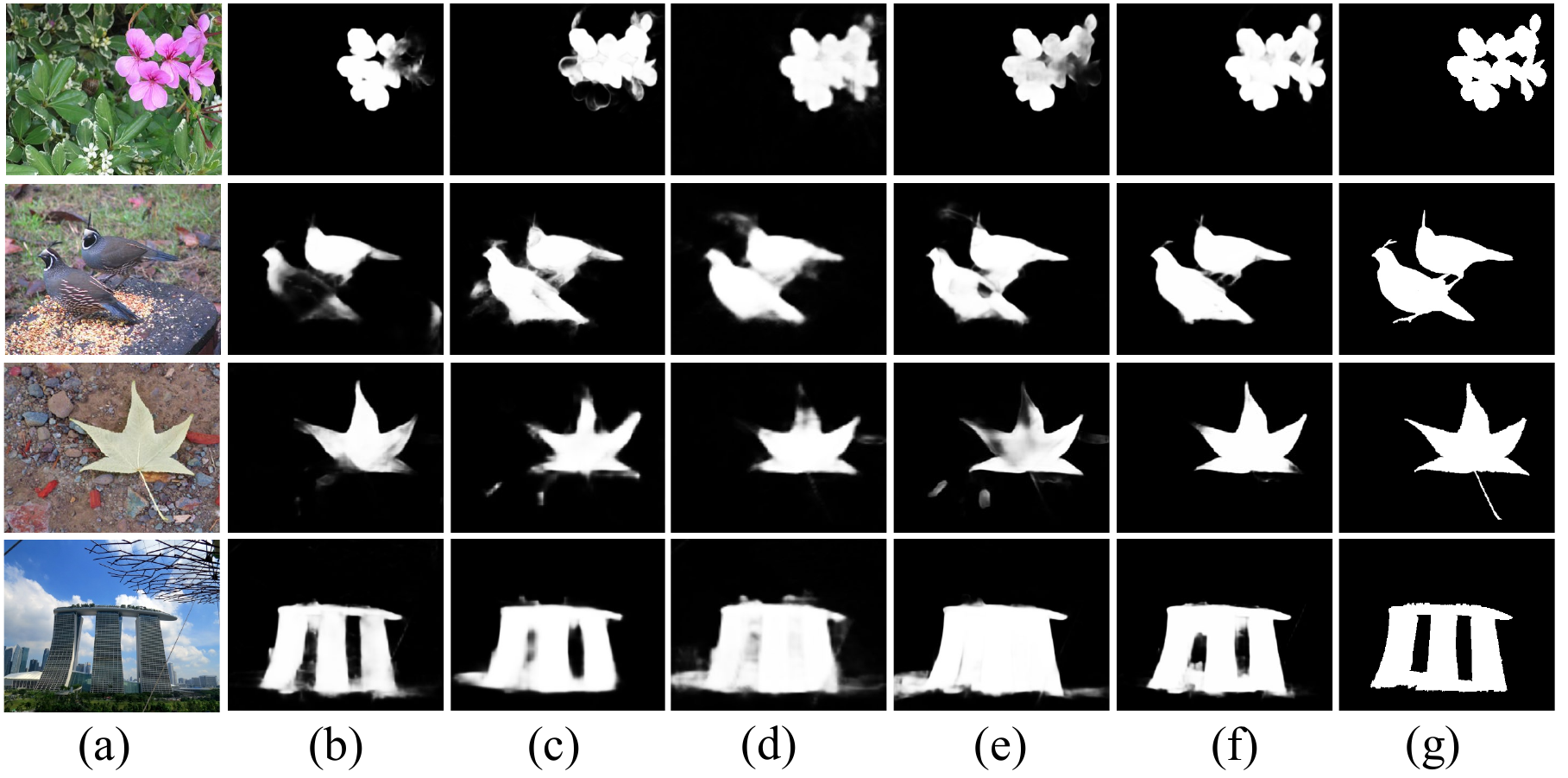}}
\vspace{-4mm}
\captionsetup{margin=1pt,justification=justified}{}
\caption{Visualization of saliency maps predicted by aggregated feature $\mathbf{F}$, FPN based models, and our method. (a) Source images. (b-f) Results of backbone + CFA, CASNet~\cite{cascaded}, PiCANet~\cite{PiCANet:liu2018picanet}, PoolNet~\cite{poolnet}, backbone + CFA + CFD. (g) Ground Truth.
}\label{fig:CFAglobalmap}
\end{center}
\vspace{-5mm}
\end{figure}

\begin{table}[t]
  \small
  \setlength\tabcolsep{0.9mm}
  \begin{tabular}{c|l|cccc} \toprule[1pt]
  \multirow{2}*{No.}&
   \multirow{2}*{Module}& \multicolumn{2}{c}{DUT-O\cite{DUT-OMRYang}} &
    \multicolumn{2}{c}{PASCAL-S\cite{PSACALSLi}}\\
    \cmidrule(l){3-4}\cmidrule(l){5-6}
    & & MaxF~$\uparrow$ & MAE~$\downarrow$ & MaxF~$\uparrow$ & MAE~$\downarrow$ \\ 
     \midrule[0.7pt]
   1& Res50 & 0.761 & 0.084 & 0.833 & 0.128 \\
   2& Res50 + FPN & 0.796 & 0.065 & 0.845 & 0.087 \\
   3& Res50 + CFA & \textbf{0.817} & \textbf{0.061} & \textbf{0.874} & \textbf{0.079} \\
   4& Res50 + CFA + CFD &\textcolor{aa}{\textbf{0.834}} & \textcolor{aa}{\textbf{0.053}} & \textcolor{aa}{\textbf{0.886}} & \textcolor{aa}{\textbf{0.072}}  \\
  \bottomrule[1pt]
  \end{tabular}
  \captionsetup{margin=1pt,justification=justified}
\caption{Ablation analysis w.r.t. effectiveness of CFA/CFD. Res50 is the ResNet-50 backbone. CFA and CFD in our method are important for improving performance. Best and second best results are shown in \textbf{black} and \textcolor{aa}{\textbf{red}}, respectively.} \label{component analyse}
\vspace{-8pt}
\end{table}

 \item \textbf{\textit{w/} CFD}:\quad Comparing Row 3 and 4 in Tab.~\ref{component analyse},
collaborating with CFD (Row 4),
the MaxF scores are improved with a margin of 1.7\% , 1.2\% on DUT-O and PASCAL-S datasets,
and the MAE values are decreased from 0.061 to 0.053 for DUT-O dataset, from 0.079 to 0.072 for PASCAL-S dataset, respectively.
Moreover,
by comparing results of FPN, 
applying  both  CFA  and  CFD  greatly improve  performance in both MaxF and MAE values.

Fig.~\ref{fig:decoder} gives the visualization feature maps at each level after CFD.
Obviously,
by comparing Rows 3 and 6 (w CFD) with Rows 1, 2, 4, and 5 (w/o CFD), 
the distributed feature maps at each fusion level provide rich semantics and clear object boundaries, 
ensuring that entire salient objects can be segmented with sharp object boundaries (Row 3 and 6 (g)).

Fig.~\ref{fig:CFAglobalmap} (f) and (b) gives Some corresponding saliency maps between w/ and w/o CFD. 
Clearly, inaccurate saliency results, e.g. over-predicted and incomplete objects,
blurred object boundaries, get greatly improved by collaborating with the CFD.
These results consistently demonstrate the effectiveness of CFD.

\end{itemize} 
\vspace{-4mm}
\begin{table}[t]
  \centering
  \small
  \setlength\tabcolsep{1.02mm}
  \begin{tabular}{c|cccccc} \toprule[1pt]
    \multirow{2}*{Module}& \multicolumn{2}{c}{DUT-O\cite{DUT-OMRYang}} &
    \multicolumn{2}{c}{PASCAL-S\cite{PSACALSLi}} &
    \multicolumn{2}{c}{DUTS-TE\cite{DUTS}} \\
	\cmidrule(l){2-3} \cmidrule(l){4-5} \cmidrule(l){6-7}
     & MaxF~$\uparrow$ & MAE~$\downarrow$ & MaxF~$\uparrow$ & MAE~$\downarrow$ & MaxF~$\uparrow$ & MAE~$\downarrow$  \\ \midrule[0.7pt]
   \bf{(A)} & 0.803 & 0.069 & 0.861 & 0.081 & 0.863& 0.049 \\
    \bf{(B)} & 0.811 & 0.064 & 0.868 & 0.078 & 0.870 & 0.047 \\
    \bf{(C)} & 0.813 & 0.062 & 0.870 & 0.079 & 0.872 & 0.047 \\
    \bf{(D)} & \textcolor{aa}{\textbf{0.817}} & \textcolor{aa}{\textbf{0.061}} & \textcolor{aa}{\textbf{0.874}} & \textcolor{aa}{\textbf{0.079}} & \textcolor{aa}{\textbf{0.875}} & \textcolor{aa}{\textbf{0.045}} \\
    \bottomrule[1pt]
  \end{tabular}
  \vspace{-2mm}
  \captionsetup{margin=1pt,justification=justified}
\caption{Ablation analysis w.r.t. different configurations of CFA.
Design of CFA achieves better performance than other settings.
Best results are shown in \textcolor{aa}{\textbf{red}}.} \label{weight analyse}
\vspace{-4mm}
\end{table}

\begin{figure*}[!t]
\begin{center}
\subfigure
{\includegraphics[width=1.0\textwidth]{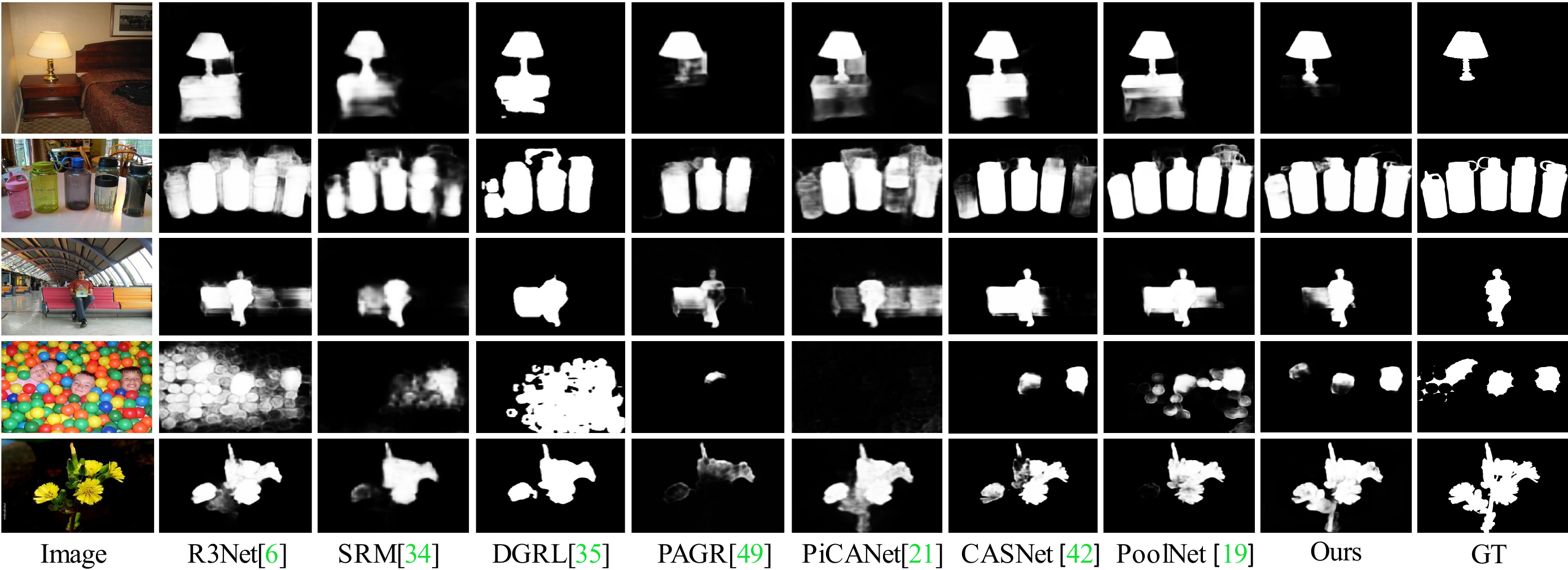}}
\vspace{-6mm}
\captionsetup{margin=8pt,justification=justified}
\caption{Comparison of saliency maps generated by our method and previous state-of-the-arts.
It can be seen that our method can not only locate the entire foreground salient objects 
but also effectively suppress cluster backgrounds, 
even for some challenging scenes. Best viewed in color.
}\label{fig:visualmap}
\end{center}
\vspace{-4mm}
\end{figure*}
\paragraph{Configurations of CFA}
We here analyze the effectiveness of our CFA design, which simultaneously considers multi-level features for adaptive layer-wise reweighting during aggregation. We compare our approach against the following baselines, including:


\begin{enumerate}[label=(\Alph*)] 
\setlength{\itemsep}{1pt}
\setlength{\parsep}{1pt}
\setlength{\parskip}{1pt}
\vspace{-0.5em}
 \item \textbf{No reweighting:}\quad The feature maps from each layer are directly concatenated, followed by a $1\times1$ conv layer for saliency map prediction.
 \item \textbf{Non-learnable reweighting:}\quad We use global average pooling (GAP) on each level features to obtain the layer-wise weights and multiply them with the original features for aggregation before producing $\mathcal{S}_{\mathrm{g}}$.
 \item \textbf{Independent layer-wise reweighting:}\quad Similar to (B), we apply GAP on each level features, followed by two fully connected layers before multiplying with the original features. This is performed \textbf{independently} on each level before concatenation.
 
 \item \textbf{Collaborative layer-wise reweighting:}\quad  We apply our CFA module to learn a set of layer-wise weights by simultaneously considering all the information among different layers for aggregation, as discussed in~\ref{CFAsubsection}.
 
\end{enumerate}
Tab.~\ref{weight analyse} reports the qualitative results of the above settings.
As can be observed,
both \textbf{(B)} and \textbf{(C)} significantly outperform the method \textbf{(A)}.
This confirms that dynamically leveraging multi-level features is crucial for saliency detection.
However, \textbf{(B)} and \textbf{(C)} give inferior performance to \textbf{(D)}, 
because the two designs reweight each level features by viewing global weights from themselves independently,
which ignores the channel interdependencies among different levels.
On the contrary, with collaborative layer-wise rewighting,
CFA obviously achieves better performance for predicting $\mathcal{S}_{\mathrm{g}}$.
These results indicate that the design of CFA 
plays an important role in boosting saliency performance.

\vspace{-3mm}
\paragraph{Configurations of CFD}
To be better illustrating the distribution process in CFD,
we allocate the aggregated feature $\mathbf{F}$ to different numbers of level features.
Tab.~\ref{CFD fusion analysis} reports the corresponding comparison results in terms of MaxF and MAE values on two challenging datasets.
By comparing results of Row 1 in Tab.~\ref{CFD fusion analysis},
the CFD module (Rows 2$\sim$6) contributes a lot to produce better saliency results.
This further demonstrates that the stage-wise fusion performs better than the single stage fusion for saliency detection.
Besides,
by distributing $\mathbf{F}$ into 1$\sim$5 levels for progressive fusion respectively,
the performance is gradually improved, illustrating that each level feature in CFD plays an important role for the progressive fusion.

\begin{table}[!t]
  \centering
  \small
  \setlength\tabcolsep{1.0mm}
  \begin{tabular}{l|l|cccc} \toprule[1pt]
  \multirow{2}*{No.}&
    \multirow{2}*{Settings}& 
    \multicolumn{2}{c}{PASCAL-S\cite{PSACALSLi}} &
    \multicolumn{2}{c}{DUTS-TE\cite{DUTS}} \\
	\cmidrule(l){3-4} \cmidrule(l){5-6} 
    & & MaxF~$\uparrow$ & MAE~$\downarrow$ & MaxF~$\uparrow$ & MAE~$\downarrow$ \\ \midrule[0.7pt]
    1&\bf{(D)} & 0.874 & 0.079 & 0.875 & 0.045 \\
    2& $\{\widetilde{\mathbf{X}}_{0}\}$ & 0.880 & 0.075 & 0.882 & 0.040 \\
    3& $\{\widetilde{\mathbf{X}}_{0}$, $\widetilde{\mathbf{X}}_{1}\}$ & 0.879 & 0.076 & 0.885 & 0.040 \\
    4& $\{\widetilde{\mathbf{X}}_{0}$, $\widetilde{\mathbf{X}}_{1}$, $\widetilde{\mathbf{X}}_{2}\}$ 
    & 0.881 & 0.074 & 0.887 & 0.038 \\
    5&\{$\widetilde{\mathbf{X}}_{0}$, $\widetilde{\mathbf{X}}_{1}$, $\widetilde{\mathbf{X}}_{2}$,
    $\widetilde{\mathbf{X}}_{3}\}$
    & 0.883 & 0.073 & 0.889 & 0.038  \\
   6& $\{\widetilde{\mathbf{X}}_{0}$, $\widetilde{\mathbf{X}}_{1}$, $\widetilde{\mathbf{X}}_{2}$,
    $\widetilde{\mathbf{X}}_{3}$,
    $\widetilde{\mathbf{X}}_{4}\}$
    & \textcolor{aa}{\textbf{0.886}} & \textcolor{aa}{\textbf{0.072}} & \textcolor{aa}{\textbf{0.896}} & \textcolor{aa}{\textbf{0.035}} \\
    \bottomrule[1pt]
  \end{tabular}
  \vspace{-2mm}
  \captionsetup{margin=1pt,justification=justified}
\caption{Ablation analysis of CFD with different distribution configurations. \textbf{(D)} refers to w/o CFD module defined in Tab.~\ref{weight analyse}.
Each level feature in CFD contributes a lot to the progressive fusion. Best results are highlighted in \textcolor{aa}{\textbf{red}}.} \label{CFD fusion analysis}
\vspace{-8pt}
\end{table}

\subsection{Comparison with State-of-the-Arts}
\label{state-of-the-art}
We compare our proposed method with
14 deep saliency detection methods, including
DCL \cite{DCL:li2016deep}, 
DSS \cite{DSS:hou2017deeply}, NLDF \cite{NLDF:luo2017non},
Amulet \cite{Amulet:zhang2017amulet}, SRM \cite{SRM:wang2017stagewise},
DGRL \cite{DGRL:wang2018detect},
R3Net \cite{R3Net:deng18r}, BMPM \cite{BMPM:zhang2018bi},
PAGR \cite{PAGR:zhang2018progressive},
PiCANet \cite{PiCANet:liu2018picanet},
AFNet \cite{AFNet},
BASNet \cite{BASNet},
CASNet \cite{cascaded},
and PoolNet \cite{poolnet}.
For fair comparison, we cite the public comparison results provided by \cite{Feng:sal_eval_toolbox}, 
which generate saliency maps from the source code released by the authors or directly provided by them.
We evaluate all the competitors with the same evaluation code.

\vspace{-3mm}
\paragraph{Visual Comparison}
Fig.~\ref{fig:visualmap} shows visual comparisons of the proposed model (Ours) with previous state-of-the-art methods.
We can clearly see that our model highlights salient objects closest to the ground-truth maps in various challenging scenarios,
including images with cluster backgrounds and foregrounds (Row 3, 4),
object having similar appearance to background (Row 1, 3, 4),
multiple instances of the same object (Row 2, 5),
and objects occluded by background objects (Row 3, 4).
More importantly,
our model can well segment the entire objects (Row 1, 2, 3, 5) with clear salient object boundaries (Row 1, 2, 3, 4, 5),
demonstrating the effectiveness of the proposed CFPN.

\begin{table*}[tp!]
  \centering
  \footnotesize
  \renewcommand{\arraystretch}{1.1}
  \renewcommand{\tabcolsep}{1.0mm}
  \begin{tabular}{l|l|cccccccccccc}
  \toprule[1pt]
  \multirow{2}*{\bf{Methods}}&\multirow{2}*{Backbone} & \multicolumn{2}{c}{ECSSD \cite{ECSSDYan}}&\multicolumn{2}{c|}{PASCAL-S \cite{PSACALSLi}}&\multicolumn{2}{c|}{DUTS-TE \cite{DUTS}}&\multicolumn{2}{c|}{HKU-IS \cite{MDF:li2015visual}}&\multicolumn{2}{c|}{SOD \cite{SODVida}}&\multicolumn{2}{c}{DUT-OMRON \cite{DUT-OMRYang}}\\
   \cmidrule(l){3-4} \cmidrule(l){5-6} \cmidrule(l){7-8} \cmidrule(l){9-10} \cmidrule(l){11-12} \cmidrule(l){13-14}
   & & MaxF~$\uparrow$ & MAE~$\downarrow$ & MaxF~$\uparrow$ & MAE~$\downarrow$ & MaxF~$\uparrow$ & MAE~$\downarrow$ & MaxF~$\uparrow$ & MAE~$\downarrow$ & MaxF~$\uparrow$ & MAE~$\downarrow$ & MaxF~$\uparrow$ & MAE~$\downarrow$ \\
  \midrule[0.8pt]
  \multicolumn{14}{l}{\textcolor{dd}{\textbf{VGG  backbone}}} \\
\midrule[0.8pt]
DCL {\tiny{CVPR2016}} ~\cite{DCL:li2016deep} & VGG-16 & 0.890 & 0.088 & 0.805 & 0.125 & 0.782 & 0.088 & 0.885 & 0.072 & 0.823 & 0.141 & 0.739 & 0.097 \\

DSS {\tiny{CVPR2016}} ~\cite{DSS:hou2017deeply} & VGG-16 & 0.916 & 0.053 & 0.836 & 0.096 & 0.825 & 0.057 & 0.911 & 0.041 & 0.844 & 0.121 & 0.771 & 0.066 \\

NLDF {\tiny{CVPR2017}} ~\cite{NLDF:luo2017non} & VGG-16 & 0.905 & 0.063 & 0.831 & 0.099 & 0.812 & 0.066 & 0.902 & 0.048 & 0.841 & 0.124 & 0.753 & 0.080 \\

Amulet {\tiny{ICCV2017}} ~\cite{Amulet:zhang2017amulet} & VGG-16 & 0.915 & 0.059 & 0.837 & 0.098 & 0.778 & 0.085 & 0.895 & 0.052 & 0.806 & 0.141 & 0.742 & 0.098 \\

BMPM {\tiny{CVPR2018}} ~\cite{BMPM:zhang2018bi} & VGG-16 & 0.929 & 0.045 & 0.862 & 0.074 & 0.851 & 0.049 & 0.921 & 0.039 & 0.855 & 0.107 & 0.774 & 0.064 \\

PAGR {\tiny{CVPR2018}} ~\cite{PAGR:zhang2018progressive} & VGG-19 & 0.927 & 0.061 & 0.856 & 0.093 & 0.855 & 0.056 & 0.918 & 0.048 & - & - & 0.771 & 0.071 \\

PiCANet {\tiny{CVPR2018}} ~\cite{PiCANet:liu2018picanet} & VGG-16 & 0.931 & 0.047 & 0.868 & 0.077 & 0.851 & 0.054 & 0.921 & 0.042 & 0.853 & 0.102 & 0.794 & 0.068 \\

AFNet {\tiny{CVPR2019}} ~\cite{AFNet} & VGG-16 & 0.935 & 0.042 & 0.868 & 0.071 & 0.862 & 0.046 & 0.923 & 0.036 & 0.856 & 0.109 & 0.797 & 0.057 \\
PoolNet {\tiny{CVPR2019}} ~\cite{poolnet} & VGG-16 & 0.936 & 0.047 & 0.857 & 0.078 & 0.876 & 0.043 & 0.928 & 0.035 & 0.859 & 0.115 & 0.817 & 0.058 \\
\hline
\textbf{Ours (VGG)} & VGG-16 & 
\textcolor{bb}{\textbf{0.943}} & \textcolor{bb}{\textbf{0.040}} & \textcolor{bb}{\textbf{0.874}} & \textcolor{bb}{\textbf{0.071}} & \textcolor{bb}{\textbf{0.885}} & \textcolor{bb}{\textbf{0.038}} & \textcolor{bb}{\textbf{0.937}} & \textcolor{bb}{\textbf{0.031}} & \textcolor{bb}{\textbf{0.870}} & \textcolor{bb}{\textbf{0.097}} & \textcolor{bb}{\textbf{0.829}} & \textcolor{bb}{\textbf{0.054}} \\[1pt]
  \midrule[0.8pt]
  \multicolumn{13}{l}{\textcolor{dd}{\textbf{ResNet  backbone}}} \\[1pt]
\noalign{\smallskip}\hline\noalign{\smallskip}
SRM {\tiny{ICCV2017}} ~\cite{SRM:wang2017stagewise} & ResNet-50 & 0.917 & 0.054 & 0.847 & 0.085 & 0.827 & 0.059 & 0.906 & 0.046& 0.843 & 0.127 & 0.769 & 0.069 \\[1pt]

DGRL {\tiny{CVPR2018}} ~\cite{DGRL:wang2018detect} & ResNet-50 & 0.922 & 0.041 & 0.854 & 0.078 & 0.829 & 0.056 & 0.910 & 0.036 & 0.845 & 0.104 & 0.774 & 0.062 \\[1pt]

R3Net {\tiny{IJCAI2018}} ~\cite{R3Net:deng18r} & ResNeXt & 0.931 & 0.046 & 0.845 & 0.097 & 0.828 & 0.059 & 0.917 & 0.038 & 0.836 & 0.136 & 0.792 & 0.061 \\[1pt]
%
PiCANet {\tiny{CVPR2018}} ~\cite{PiCANet:liu2018picanet} & ResNet-50 & 0.935 & 0.047 & 0.881 & 0.087 & 0.860 & 0.051 & 0.919 & 0.043 & 0.858 & 0.109 & 0.803 & 0.065 \\[1pt]
%
BASNet {\tiny{CVPR2019}} ~\cite{poolnet} & ResNet-34 & 0.942 & 0.037 & 0.854  & 0.076 & 0.860 & 0.047 & 0.928 & 0.032 & 0.851 & 0.114 & 0.805 & 0.056\\[1pt]
%
CASNet {\tiny{CVPR2019}} ~\cite{cascaded} & ResNet-50 & 0.939 & 0.037 & 0.864  & 0.072 & 0.865 & 0.043 & 0.925 & 0.034 & - & - & 0.797 & 0.056\\[1pt]
%
PoolNet {\tiny{CVPR2019}} ~\cite{poolnet} & ResNet-50 & 0.940 & 0.042 & 0.863  & 0.075 & 0.886 & 0.040 & 0.934 & 0.032 & 0.867 & 0.100 & 0.830 & 0.055\\[1pt]
\hline
\textbf{Ours (Res18)} & ResNet-18 &  \textbf{0.942} & \textbf{0.039} & \textbf{0.879} & \textbf{0.074} & \textbf{0.887} & \textbf{0.039} & \textbf{0.933} & \textbf{0.032} & \textbf{0.872} & \textbf{0.085} & \textbf{0.821} & \textbf{0.055} \\[1pt]
\textbf{Ours (Res50)}  & ResNet-50 & \textcolor{aa}{\textbf{0.948}} & \textcolor{aa}{\textbf{0.035}} &  \textcolor{aa}{\textbf{0.886}} & \textcolor{aa}{\textbf{0.072}} & \textcolor{aa}{\textbf{0.896}} & \textcolor{aa}{\textbf{0.035}} & \textcolor{aa}{\textbf{0.940}} & \textcolor{aa}{\textbf{0.029}} & \textcolor{aa}{\textbf{0.873}} & \textcolor{aa}{\textbf{0.083}} & \textcolor{aa}{\textbf{0.834}} & \textcolor{aa}{\textbf{0.053}} \\
%
\bottomrule[1pt]
\end{tabular}
\captionsetup{margin=8pt,justification=justified}
\caption{Comparisons of max F-measure and MAE values on VGG \cite{VGG} and ResNet~\cite{resnet:He2015Deep} backbones are reported. 
Results of our method are shown in \textcolor{bb}{\textbf{blue}}, \textbf{black}, and \textcolor{aa}{\textbf{red}}, respectively. 
With different backbones,
the proposed method consistently achieves better performance than the previous state-of-the-arts. Best viewed in color.}\label{allCompare}
\vspace{-5mm}
\end{table*}

\begin{figure*}[!t]
\begin{center}
\subfigure
{\includegraphics[width=0.33\textwidth]{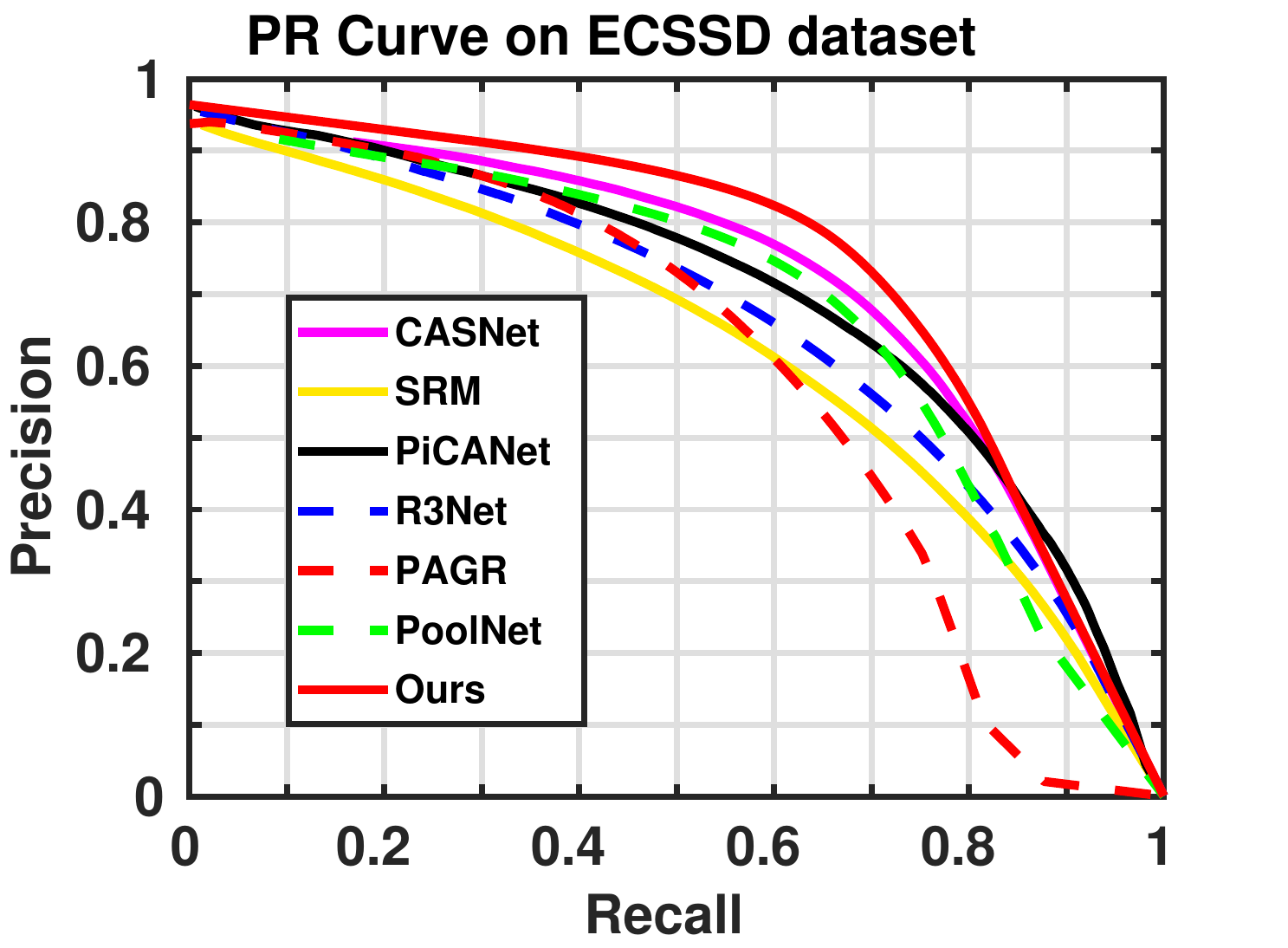}}
\subfigure
{\includegraphics[width=0.33\textwidth]{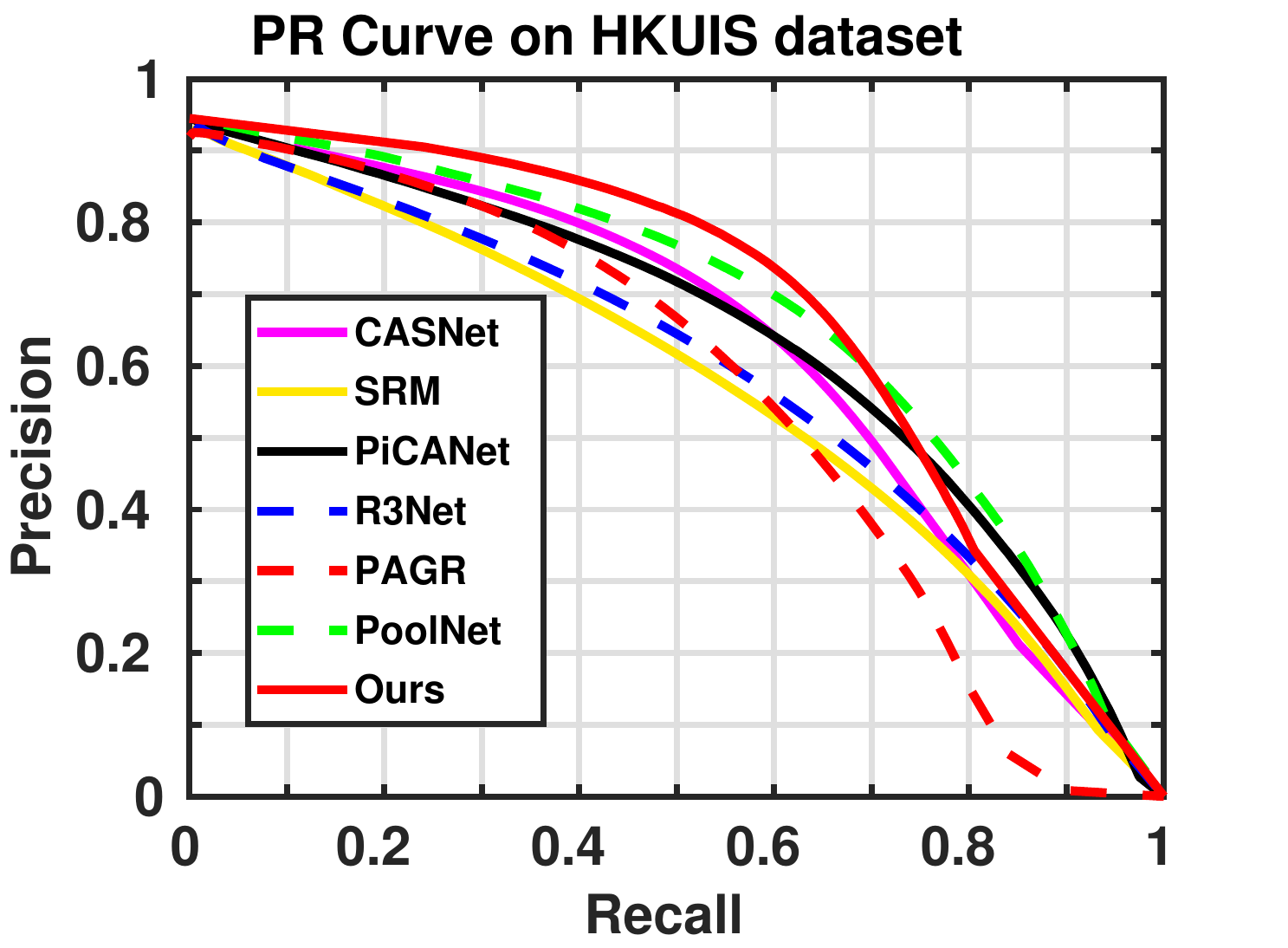}}
\subfigure
{\includegraphics[width=0.33\textwidth]{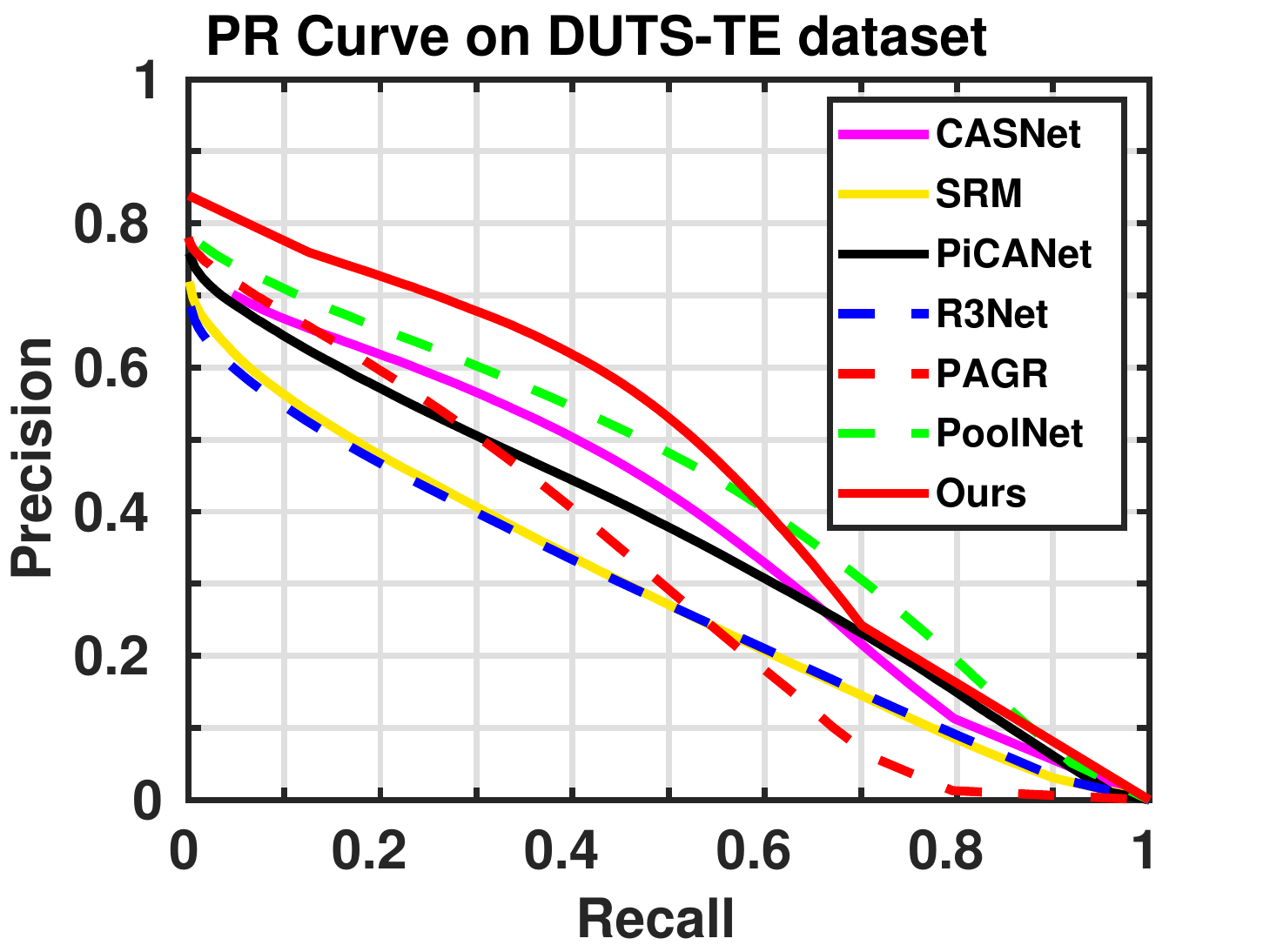}}
\vspace{-6mm}
\captionsetup{margin=8pt,justification=justified}
\caption{Precision and recall curves on ECSSD~\cite{ECSSDYan}, HKUIS~\cite{HKU-ISLi}, and DUTS-TE~\cite{DUTS} datasets.
The proposed method outperforms previous state-of-the-arts on all the datasets. Best viewed in color.}\label{fig:allPR}
\end{center}
\vspace{-5mm}
\end{figure*}

\vspace{-4mm}
\paragraph{F-measure and MAE Comparison}
Tab. \ref{allCompare} reports the MaxF and MAE scores of our method using different backbones (VGG-16 \cite{VGG}, ResNet-18 \cite{resnet:He2015Deep}, and ResNet-50 \cite{resnet:He2015Deep}) compared with other methods.
Obviously, CFPN achieves excellent results on all the datasets with the similar backbones across the metrics.
In particular,
with both VGG-16 \cite{VGG} and ResNet-50 \cite{resnet:He2015Deep} backbones,
CFPN shows significantly improved $F_{\beta}$-max scores compared with the second best PoolNet~\cite{poolnet}, on the more challenging benchmarks PASCAL-S (\textbf{VGG-16}: 0.874 \textit{vs} 0.857; \textbf{ResNet-50}: 0.886 \textit{vs} 0.863), DUTS-TE  (\textbf{VGG-16}: 0.885 \textit{vs} 0.876; \textbf{ResNet-50}: 0.896 \textit{vs} 0.886), and HKUIS (\textbf{VGG-16}: 0.937 \textit{vs} 0.928; \textbf{ResNet-50}: 0.940 \textit{vs} 0.934). 
More importantly,
when using ResNet-18~\cite{resnet:He2015Deep} as backbone,
our CFPN not only outperforms all the previous VGG backbone approaches significantly, but also beats most of the ResNet-50 based methods, especially on the more challenging datasets including  PASCAL-S, SOD, and DUTS-TE.
These results clearly illustrate the superior performance and robustness of CFPN.

\vspace{-3mm}
\paragraph{PR Curves Comparison}
We also give the precision-recall curves in Fig. \ref{fig:allPR}.
Due to limited space,
we simply show the PR curves of the previous methods implemented with ResNet-50 backbone over three widely used datasets.
As can be seen, the PR curves of our CFPN, represented by the straight 
red lines, consistently outperform all other previous models over all datasets.
These results convincingly demonstrate the effectiveness of our method.

\section{Conclusion}
In this paper,
we identify the limitation of FPN based saleincy methods (i.e., \emph{indirect} information propagation between deeper and shallower layers)
and presented a novel architecture, CFPN, for salient object detection.
It consists of two essential modules: a cross-layer feature aggregation module and a cross-layer feature distribution module.
Benefiting from these two collaborative modules,
efficient information communication across multiple layers is conducted,
which reduces the information loss during FPN stage-wise fusion,
and thus leads to more accurate saliency results.
Comprehensive experiments on popular saliency detection benchmarks demonstrate the effectiveness and robustness of the proposed CFPN.
{\small
\bibliographystyle{ieee_fullname}
\bibliography{egbib}
}
\end{document}